\documentclass[10pt,twocolumn,letterpaper]{article}

\usepackage{cvpr}
\usepackage{times}
\usepackage{epsfig}
\usepackage{graphicx}
\usepackage{amsmath}
\usepackage{amssymb}
\usepackage{multirow}

\usepackage{wasysym} 
\usepackage[normalem]{ulem}
\usepackage{subcaption}
\usepackage{ctable}

\usepackage[pagebackref=true,breaklinks=true,letterpaper=true,colorlinks,bookmarks=false]{hyperref}

\cvprfinalcopy 

\def\cvprPaperID{7410} 

\ifcvprfinal\fi

\begin{document}

\title{Flow-Distilled IP Two-Stream Networks for Compressed Video Action Recognition}

\author{Shiyuan Huang
\and
Xudong Lin
\and
Svebor Karaman
\and 
Shih-Fu Chang
\and
\\
Columbia University \hspace{30pt} 
}

\maketitle

\begin{abstract}
Two-stream networks have achieved great success in video recognition. A two-stream network combines a spatial stream of RGB frames and a temporal stream of Optical Flow to make predictions. However, the temporal redundancy of RGB frames as well as the high-cost of optical flow computation creates challenges for both the performance and efficiency. Recent works instead use modern compressed video modalities as an alternative to the RGB spatial stream and improve the inference speed by orders of magnitudes. Previous works create one stream for each modality which are combined with an additional temporal stream through late fusion. This is redundant since some modalities like motion vectors
already contain temporal information. Based on this observation, we propose a compressed domain two-stream network (\textbf{IP TSN}) for compressed video recognition, where the two streams are represented by the two types of frames (I and P frames) in compressed videos, without needing a separate temporal stream. With this goal, we propose to fully exploit the motion information of P-stream through generalized distillation from optical flow, which largely improves the efficiency and accuracy. Our P-stream runs 60 times faster than using optical flow while achieving higher accuracy. Our full \textbf{IP TSN}, evaluated over public action recognition benchmarks (UCF101, HMDB51 and a subset of Kinetics), outperforms other compressed domain methods by large margins while improving the total inference speed by 20\%.  
\end{abstract}

\section{Introduction}

\begin{figure}[tbh]
\includegraphics[width=\linewidth]{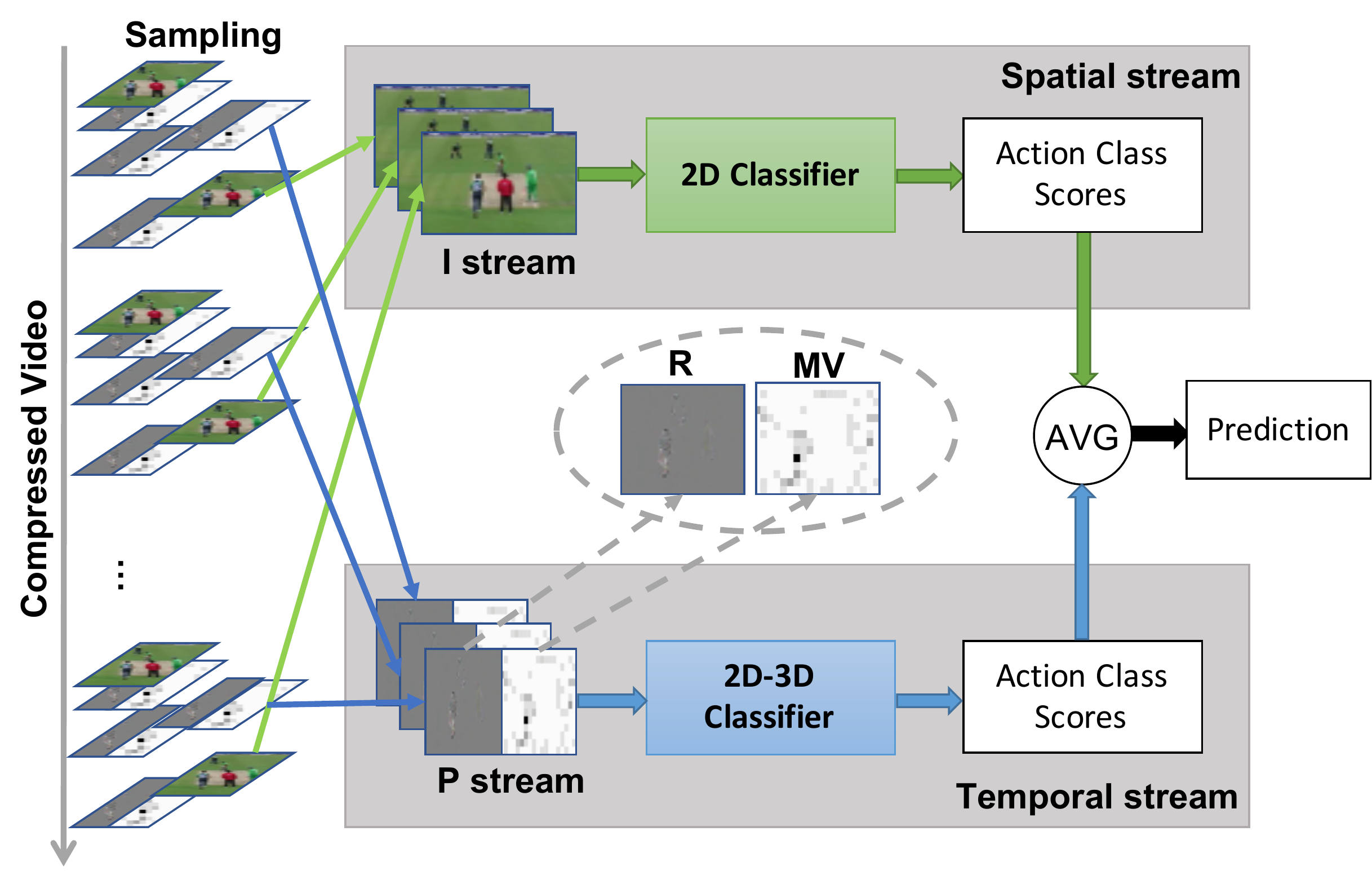}
\caption{Overview of our IP Two-Stream Network (\textbf{IP TSN}). We propose a novel compressed domain two-stream network, which exploits the two frame types readily available in compressed videos. 
The I-stream samples RGB I-frames and process them with a 2D-Net to get still scene information.
The P-stream samples the weaker P-frame modality, motion vectors (\textit{MV}) and residual errors (\textit{R}), processed by a 2D-3D Net for effective temporal modeling. The two streams outputs are fused to obtain the final prediction. Our model achieves significant performance gains in both accuracy and efficiency.}
\label{fig:ip_tsn}
\end{figure}

Video is nowadays the dominating visual data source in all kinds of application scenarios. 
The temporal context embedded in videos 
captures additional information compared to still images. 
However, the temporal aspect of videos also induces information redundancy and computational burden. 
Early works~\cite{KarpathyCVPR14,tran2017convnet} on video recognition apply convolutional neural networks (CNNs) directly on RGB frame sequences decoded from video but get limited success, due to the difficulty of extracting discriminative spatiotemporal information from the frames that highly resemble each other.
Recent works~\cite{TSN,feichtenhofer2016convolutional} have demonstrated the effectiveness of two-stream networks where appearance and motion are modeled by separate CNNs and then fused together to obtain the final prediction. 
Most state-of-the-art works keep RGB frame sequences as the appearance modality, and use optical flow (\textit{OF}) \cite{TVL1} calculated from consecutive RGB frames as the motion modality. 
Though adding an optical flow stream improves the performance, it is known to be computationally heavy. 
Some works \cite{Sun2018PWC-Net,ilg2017flownet} try to estimate OF using CNNs. But to make pixel-level accuracy, most of those networks are heavy. OF estimation remains a huge computational bottleneck for two stream frameworks.

To improve video recognition efficiency, some recent works have started to look into modern compressed video encoding formats (MPEG4, H.264, etc) as input. 
CoViAR \cite{wu2018coviar} and DMC-Net \cite{Shou_2019_CVPR} consider MPEG4 \cite{le1991mpeg} videos that contain I-frame (intra-coded frames as RGB images) and P-frame (predictive frames) represented by both Motion Vectors (\textit{MV}) and Residuals (\textit{R}). 
To apply a two-stream framework on compressed videos, the spatial stream is replaced by three separate 2D CNNs for I-frames, MV and R of P-frames respectively, and fuse all three with another temporal stream.
However, \textit{MV}s contain motion information rather than appearance information, and Rs contain rich motion boundary information~\cite{gcpr18, wu2018coviar}. 
Thus, we argue that both modalities are 
motion related
and hence not appropriate appearance modalities and that 
a proper compressed-domain two-stream framework should take this property into account when dealing with a compressed video as input.
   
DMC-Net \cite{Shou_2019_CVPR} proposes to make the traditional two-stream framework compressed-domain exclusive by replacing OF with a fast ``Discriminative Motion Cue'' generated from \textit{MV} and \textit{R}. 
Without the expensive OF calculation, DMC-Net is able to achieve a speed up of orders of magnitudes over OF-based two-stream works.  
DMC-Net uses a lightweight DenseNet generator to generate a OF-like motion representation at full-resolution. 
The network is trained using GAN loss supervised by OF and classification loss. 
However, training with GAN loss is known to be unstable, and to exploit motion information from \textit{MV} and R, explicit reconstruction of OF is not necessary. 
Several works~\cite{gcpr18, ng2018actionflownet} have demonstrated that the End-Point-Error (EPE) accuracy of OF does not guarantee recognition accuracy. 
\cite{ZhangWWQW16_MVCNN} also demonstrates the potential of \textit{MV} through soft label distillation from OF. 
Hence our insight is that we can directly guide \textit{MV} and \textit{R} to mimic OF without explicit reconstruction.
 
Based on the above observations of current compressed domain methods, we propose a new compressed-domain specific two-stream network, IP two-stream network (\textbf{IP TSN}), depicted in Figure~\ref{fig:ip_tsn}, where the spatial stream (I-stream) is modeled by RGB I-frame only and the temporal stream (P-stream) is modeled by a unified network for \textit{MV} and \textit{R}. 
Based on the nature of video encoding format which consists of detailed RGB I-frame and 
weaker
P-frame, we use a strong 2D CNN in the spatial stream to extract still scene information from sparse I-frames, and a light 2D-3D CNN in the temporal stream for better spatio-temporal modeling from P-frames. 
We also propose to exploit high-level feature supervision from OF, during training time only, to better extract motion information from \textit{MV} and \textit{R}.

Compared to other compressed-domain methods, we are able to reduce the total inference GFLOPS by $20 \%$ while increasing the accuracy by $\sim7\%$ on HMDB-51 and $\sim3\%$ on UCF-101. 

Our main contributions are:
\begin{enumerate}
\item We propose a novel compressed-domain two-stream framework exploiting the unique compressed video structure. It is able to achieve higher accuracy than other compressed-domain methods by large margins while improving the efficiency significantly.  
\item We propose to apply generalized distillation techniques to better exploit motion information from components of compressed videos, namely, \textit{MV} and \textit{R}. With the supervision of high-level optical flow (\textit{OF}) features during training time only, we are able to replace OF with \textit{MV} and \textit{R}. It speeds up the temporal stream by more than 60 times compared to using optical flow while achieving similar or better accuracy. 
\item By evaluating our methods on public action recognition benchmarks, HMDB-51, UCF-101, and a subset of Kinetics, we demonstrated the effectiveness of our framework. We are able to make large improvements over current compressed-domain methods and get significantly closer to the upper-bound performance using conventional decoded videos.

\end{enumerate}

\section{Related Works}


\noindent \textbf{Two Stream Networks for Video Action Recognition.} 
Two stream networks are first proposed by~\cite{feichtenhofer2016convolutional,Simonyan14b} where separate 2D ConvNets are used for RGB frames and OF, 
and fuse their scores for final prediction. 
Later, 3D ConvNets~\cite{Kinetics, tran2017convnet, xie2017rethinking} are shown to perform better in spatiotemporal modeling. 
With many large-scale video datasets \cite{gu2017ava,Damen2018EPICKITCHENS,activitynet,THUMOS15} coming out, 3D ConvNets are able to get high accuracies when incorporated into a two-stream framework. 
However, 3D ConvNets are computationally heavy and there are some efforts like~\cite{Tran_2018_CVPR,Tran_2019_ICCV},
trying to alleviate 3D computations and get comparable or even better performances.  
\cite{Zolfaghari_2018_ECCV} and \cite{tran2018closer} partially insert 3D layers into the network arguing that temporal modeling is only needed at certain stages. 
Based on these works, we decide to adopt a few 3D layers in our framework to enhance the temporal modeling of videos while keeping the computational cost as low as possible. 

\begin{figure}[tb]
     \centering
     \begin{subfigure}[h]{\linewidth}
         \centering
         \includegraphics[width=\linewidth]{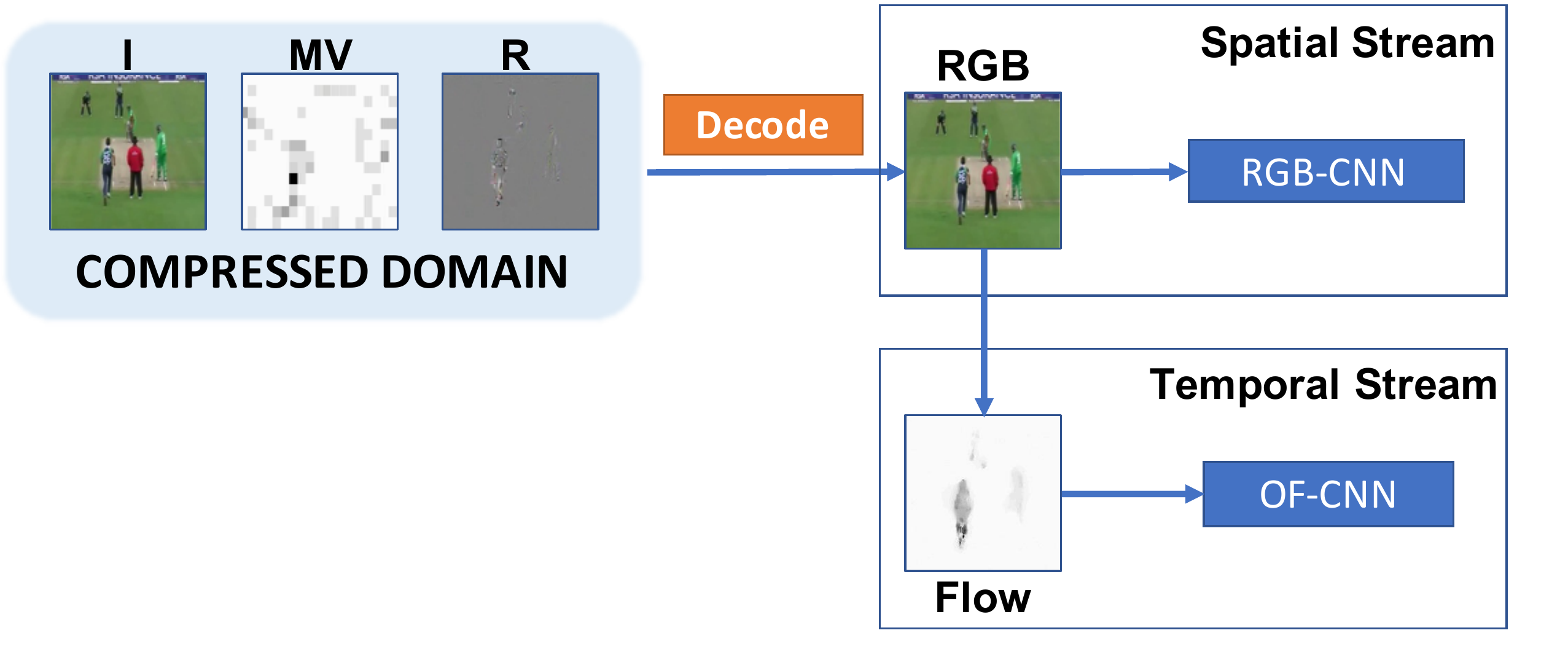}
         \caption{Two-Stream Network with Decoded Video}
         \label{fig:tsn_decode}
     \end{subfigure}
     \hfill
     \begin{subfigure}[h]{\linewidth}
         \centering
         \includegraphics[width=\linewidth]{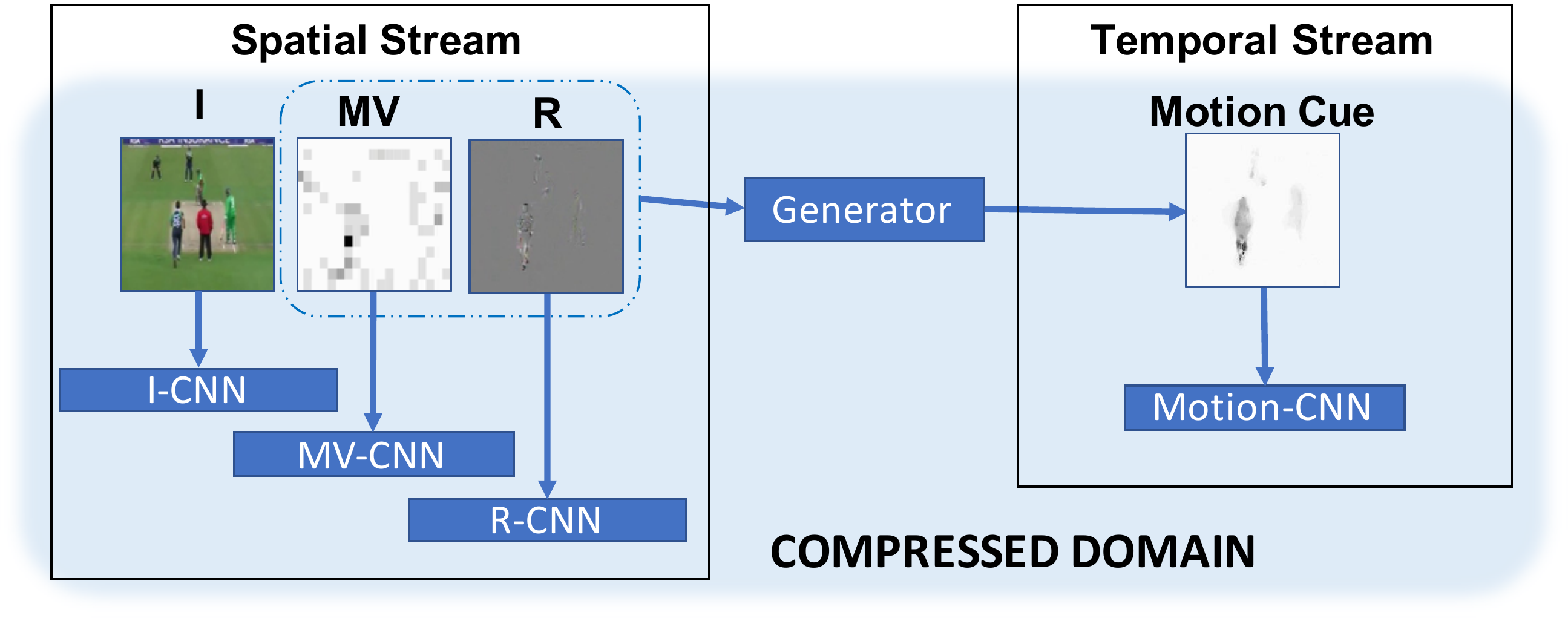}
         \caption{DMC-Net}
         \label{fig:tsn_dmcnet}
     \end{subfigure}
     \hfill
     \begin{subfigure}[h]{0.8\linewidth}
         \centering
         \includegraphics[width=\linewidth]{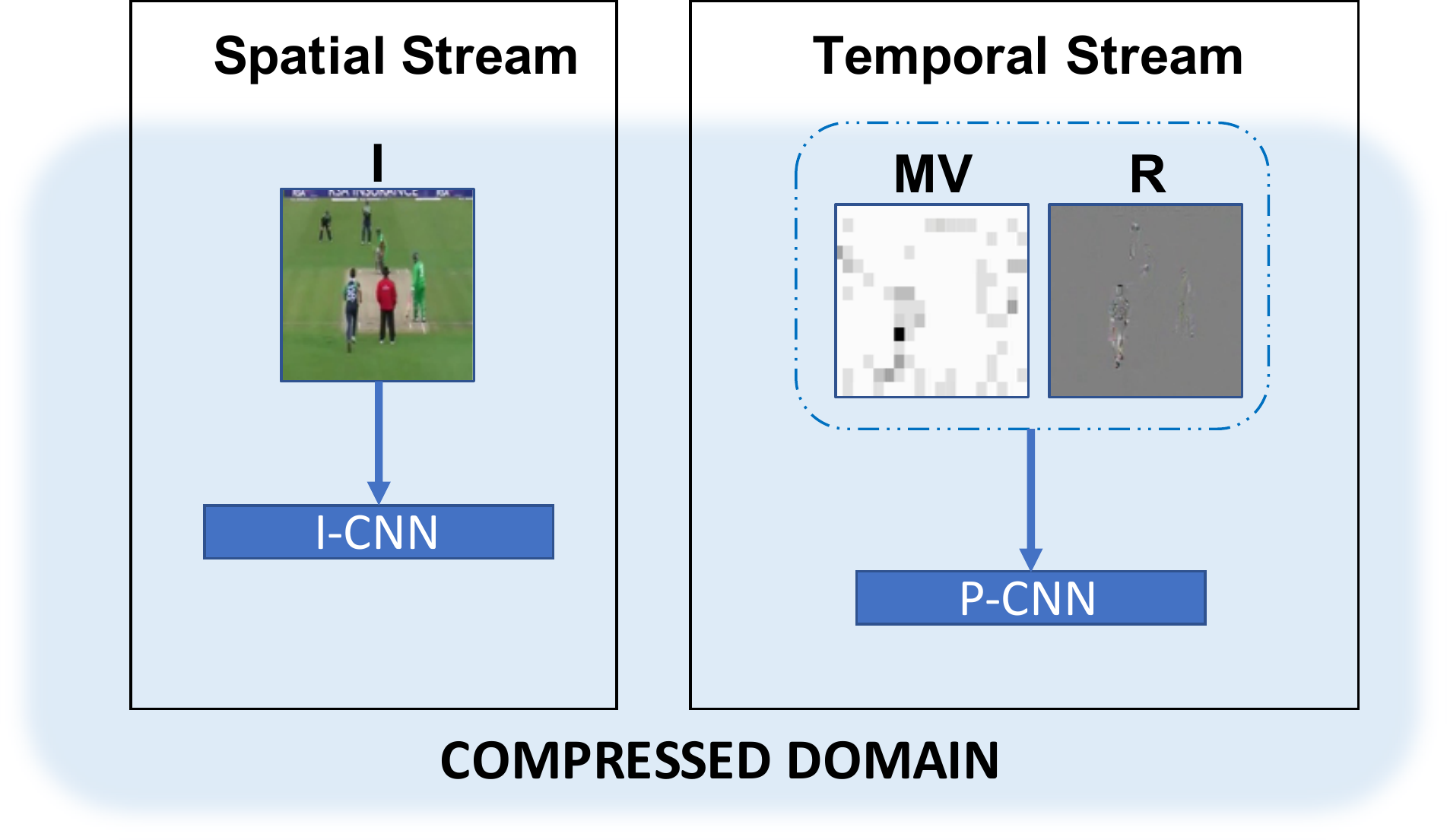}
         \caption{IP Two-Stream Network (ours)}
         \label{fig:tsn_iptsn}
     \end{subfigure}
        \caption{Two-stream networks in decoded and compressed video domains. (a) a traditional TSN \cite{Simonyan14b} taking decoded RGB and pre-computed \textit{OF} as input, (b) DMC-Net operates in the compressed domain only by merging three modality streams with a motion stream which input is computed from \textit{MV} and \textit{R}, (c) our \textbf{IP TSN} also operates solely in the compressed domain but takes directly \textit{I} and \textit{MV}-\textit{R} as the spatial and temporal stream input respectively. We do not need extra computation, e.g. the generator in DMC-Net, to get the motion modality.}
        \label{fig:tsn}
\end{figure}

All the two-stream networks discuss above in decoded videos requires additional calculation of OF and takes both RGB and OF as inputs, see Figure~\ref{fig:tsn_decode}. 
Each RGB frame and its OF frame are temporally aligned. In contrast, our IP TSN simply use two different compressed video frame types as streams inputs that are not temporally aligned. 
All the inputs are readily available in the compressed video. 
Also, given the different densities of stream inputs, we use different networks for \textit{I} and \textit{P} streams.


\noindent \textbf{Compressed Video Action Recognition.}
Recent works have started to look into compressed video domain and showed that compressed video modalities contain rich information that can be exploited quickly using light-weight networks. \cite{ZhangWWQW16_MVCNN} utilizes motion vector as a cheap alternative to optical flow, but it still needs the full decoded RGB stream in the traditional two stream setting. CoViAR~\cite{wu2018coviar} is the first work that avoids any decoding and only exploits compressed video data, i.e., I-frame, motion vectors, and residuals. It uses three separate networks for RGB I-frame, motion vectors and residuals, and gets high efficiency when averaging the per-frame inference time of all three modalities. However, in order to get comparable performance with traditional two-stream methods of decoded videos, it has to add back an optical flow stream which does require decoding and is well known to be computationally expensive. 
To solve this issue, DMC-Net \cite{Shou_2019_CVPR} proposes a fast discriminative motion representation directly learned from \textit{MV} and \textit{R} to replace the \textit{OF} stream as illustrated in Figure~\ref{fig:tsn_dmcnet}. It uses a lightweight generator to get a full resolution representation that is close to OF while jointly training with classification loss. 
However, during test time, to achieve good accuracy, both CoViAR and DMC need to sample 25 instances for each modality during test time fed to separate networks. (3-stream for CoViAR and 4-stream for DMC). 
Instead, we propose a compressed domain two-stream network that requires less sampling instances and much less computations during test time while largely improving the performance from other compressed video recognition methods. 
    
\noindent \textbf{Generalized Distillation.}
Knowledge distillation was first proposed by \cite{hinton2015distilling} as a concept of transferring knowledge from a high-performance complex model (teacher) to a simple model (student) through the supervision of complex model soft predictions. 
Recent works \cite{gupta2016cross,luo2018graph,garcia2019learning} apply the concept along with privileged information \cite{lopez_distill} in cross-modality transfer learning and show promising results.  \cite{crasto2019mars, stroud2018d3d} propose to use knowledge distilled from the optical flow to get motion representations directly from RGB inputs. \cite{ZhangWWQW16_MVCNN,Zhang2018RealTimeAR} applies similar techniques but on the motion vectors. 
All these works either consider only one student modality, or multiple student modalities independently. Our work takes optical flow as the teacher to enhance the P stream in a similar spirit. However, we consider a new student modality, P frame, which by itself is multi-modal since it consists of both \textit{MV} and \textit{R.} 
Our insight is that \textit{MV} and \textit{R} contain complementary motion information that are well aligned, and hence can be merged into one stream and supervised by \textit{OF}.


\begin{figure*}[tb]
\includegraphics[width=\linewidth]{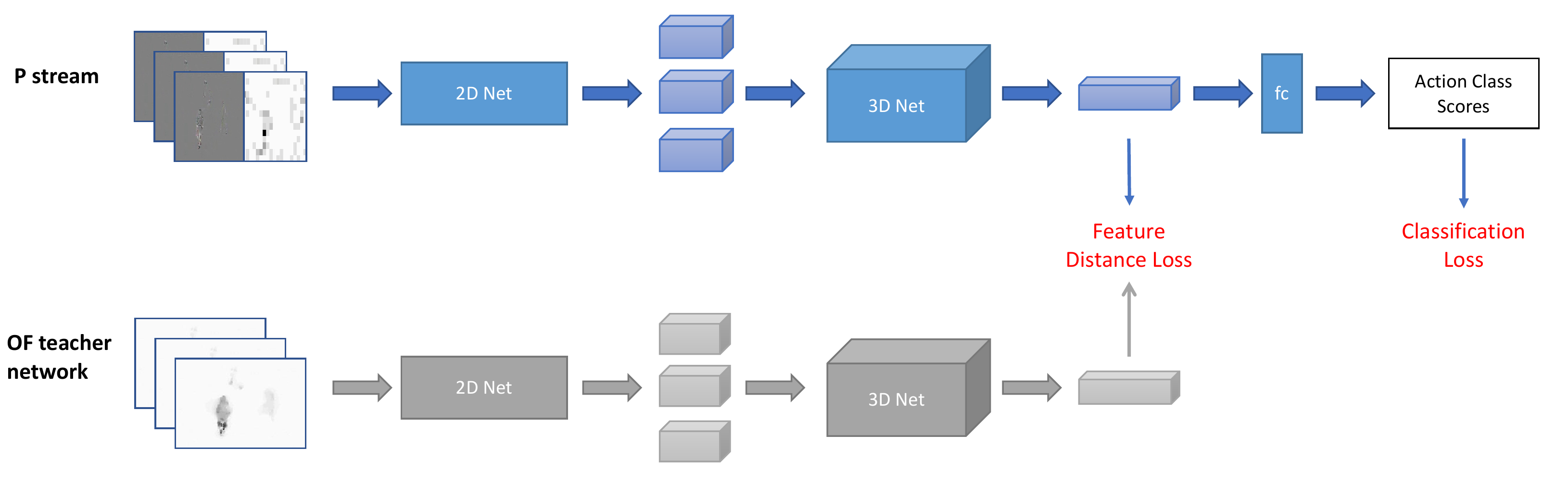}
\caption{Details of our P-stream network.  We use a  2D-3D Res18 that partially inflates a ResNet18~\cite{He2015DeepRL}. The P-stream takes as input  a clip of stacked \textit{MV} and \textit{R} under supervision of ground truth action labels. Furthermore, during training, we exploit a pre-trained \textit{OF} net as teacher to distill its \textit{OF} based features. At inference time, only the P-stream (blue parts) remains. }
\label{fig:pstream}
\end{figure*}

\section{Approach}
In this section we first define the compressed video notations. 
Then we present our \textbf{IP TSN} for compressed video recognition. 
Finally, we introduce the distillation technique we use to enhance P-stream.

\subsection{Compressed Video Formats and Notations}
Following CoViAR \cite{wu2018coviar} and DMC-Net \cite{Shou_2019_CVPR}, we consider MPEG-4 Part2 \cite{le1991mpeg} encoded videos with Group of Structure (GOP) size 12.
Starting with the first frame, I-frames appear every 12 frames, with 11 P-frames filling in between. 
Each I-frame is a full resolution RGB image (\textit{I}). 
Each P-frame is represented by a Motion Vector (\textit{MV}) computed from $16\times16$ macroblock displacement from the previous frame and a Residual (\textit{R}) computed from the RGB difference between the original image and motion compensated image. 
Both \textit{I} and \textit{R} are three channel (RGB) images of the same resolution as in the original video, while \textit{MV} is two channel of horizontal and vertical displacements that has $16\times$ smaller resolution in effect. 
In visualization, \textit{MV}s are blocky images where each $16\times16$ block is filled with the same value of its displacement. 
Additionally, we denote the optical flow as \textit{OF}, which is the pixel displacement calculated from RGB frames. In our experiment, we use TV-L1~\cite{TVL1} to extract optical flows. We use the FFmpeg~\cite{FFmpeg} based implementation provided by CoViAR~\cite{wu2018coviar} to extract \textit{I}, \textit{MV} and \textit{R} from MPEG4 videos.


\subsection{IP TSN for Compressed Video Recognition}
\label{sec:ip_tsn}

Most two-stream methods use decoded RGB frames and \textit{OF} as spatial and temporal modalities. 
Existing compressed domain methods use all three compressed video modalities, \textit{I}, \textit{MV} and \textit{R}, as the spatial stream, potentially fused with another temporal stream.
As a result, there are in total 4 separate networks adapted for 4 different modalities. 
Also, I-stream uses a strong 2D CNN as the feature extractor, while \textit{MV} and \textit{R} are considered weak modalities and hence are processed by lightweight 2D CNNs. 
    
Existing compressed domain framework are limited as \textit{MV} is computed from block displacement, which implies motions, and \textit{R} emphasizes motion boundaries, neither is appropriate as spatial modality. Rather, \textit{MV} and \textit{R} 
should more naturally be exploited as the temporal stream. 
DMC-Net~\cite{Shou_2019_CVPR} tries to reconstruct OF-like motion cues as the temporal stream. 
However, the architecture does not fully exploit the information in \textit{MV} and \textit{R} and the performance improvement is limited.

Unlike the above setting, we propose a compressed domain two-stream framework \textbf{IP TSN}, illustrated in Figure~\ref{fig:tsn_iptsn}, which is composed of an I-stream which uses \textit{I} for appearance modeling, and a P-stream which fuses \textit{MV} and \textit{R} for motion modeling. Final prediction is computed from late fusion of these two streams. 
We split a video into different frame types and feed them into separate networks for the spatial and temporal modeling purposes. Following TSN style \cite{TSN} sampling as in DMC-Net \cite{Shou_2019_CVPR} and CoViAR \cite{wu2018coviar}, we split the video into N segments sample one I-frame and one P-frame each segment, as is shown in Figure \ref{fig:ip_tsn}.

Based on the properties of the compressed video, we propose to use different network architectures for the I-stream and P-stream. 
Since I-frames have fine details, we use a heavy 2D CNN, ResNet152, as is used in \cite{wu2018coviar} to extract still scene information. 
On the other hand, \textit{MV} and \textit{R} of P-frame are weaker and noisy modalities that contain rich motion information. Recent works \cite{Zolfaghari_2018_ECCV, tran2018closer} have shown 3D CNNs at the top layers help extract high level motion features. 
Hence we use a combine 2D-3D Net for efficient motion modeling. Figure~\ref{fig:pstream} illustrates the details of the P-stream. For better comparison and efficiency, we use a 2D-3D ResNet18 where the later half of the layers are inflated into 3D. Details are given in Section~\ref{sec:experiments},  where we also study how the architectures of I and P streams affect the performance.


\begin{figure*}
    \centering
    \begin{subfigure}[h]{0.315\linewidth}
        \centering
        \includegraphics[width=\linewidth]{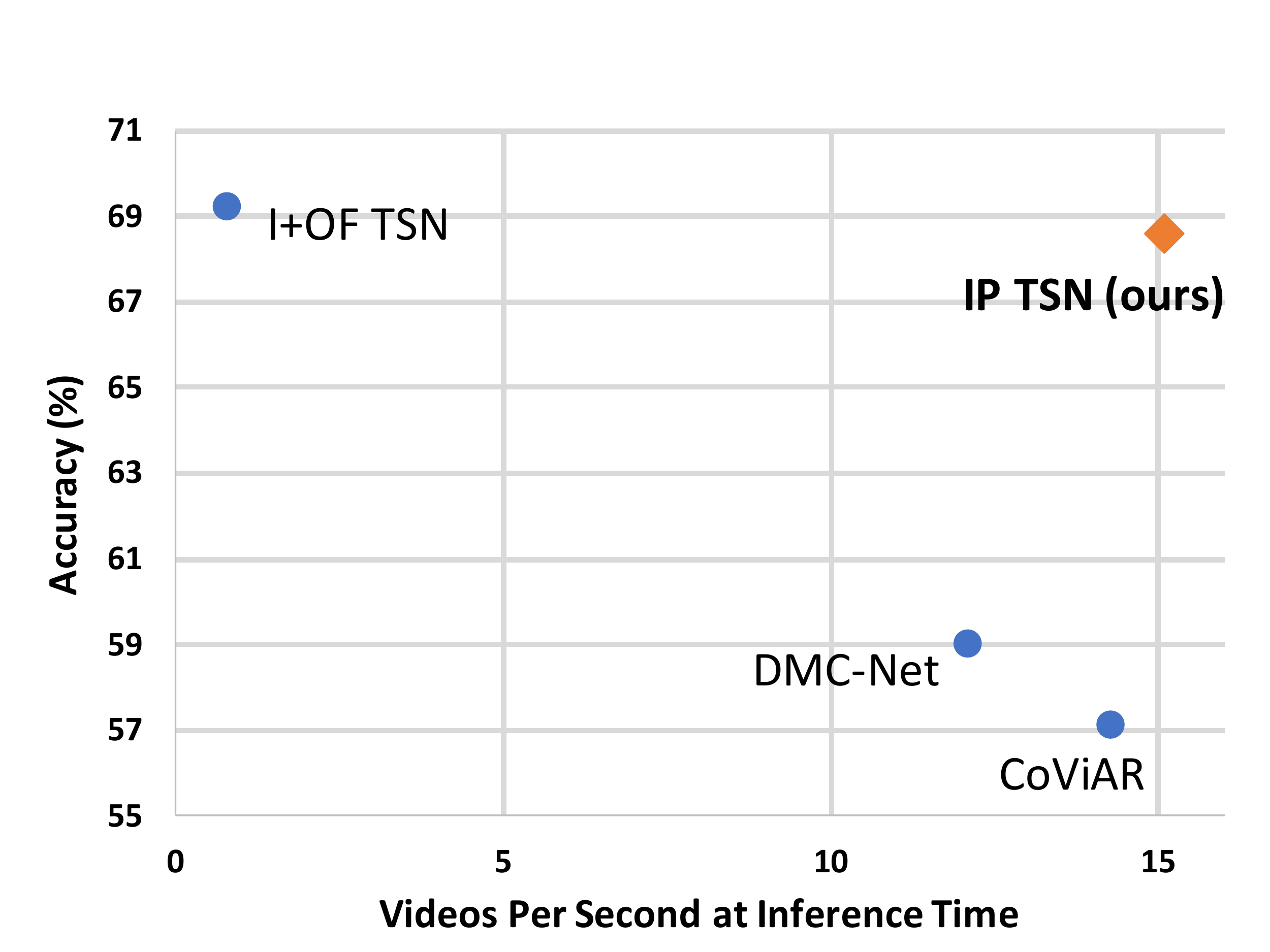}
        \caption{HMDB51}
        \label{fig:acc_hmdb}
    \end{subfigure}
    \hfill
    \begin{subfigure}[h]{0.325\linewidth}
        \centering
        \includegraphics[width=\linewidth]{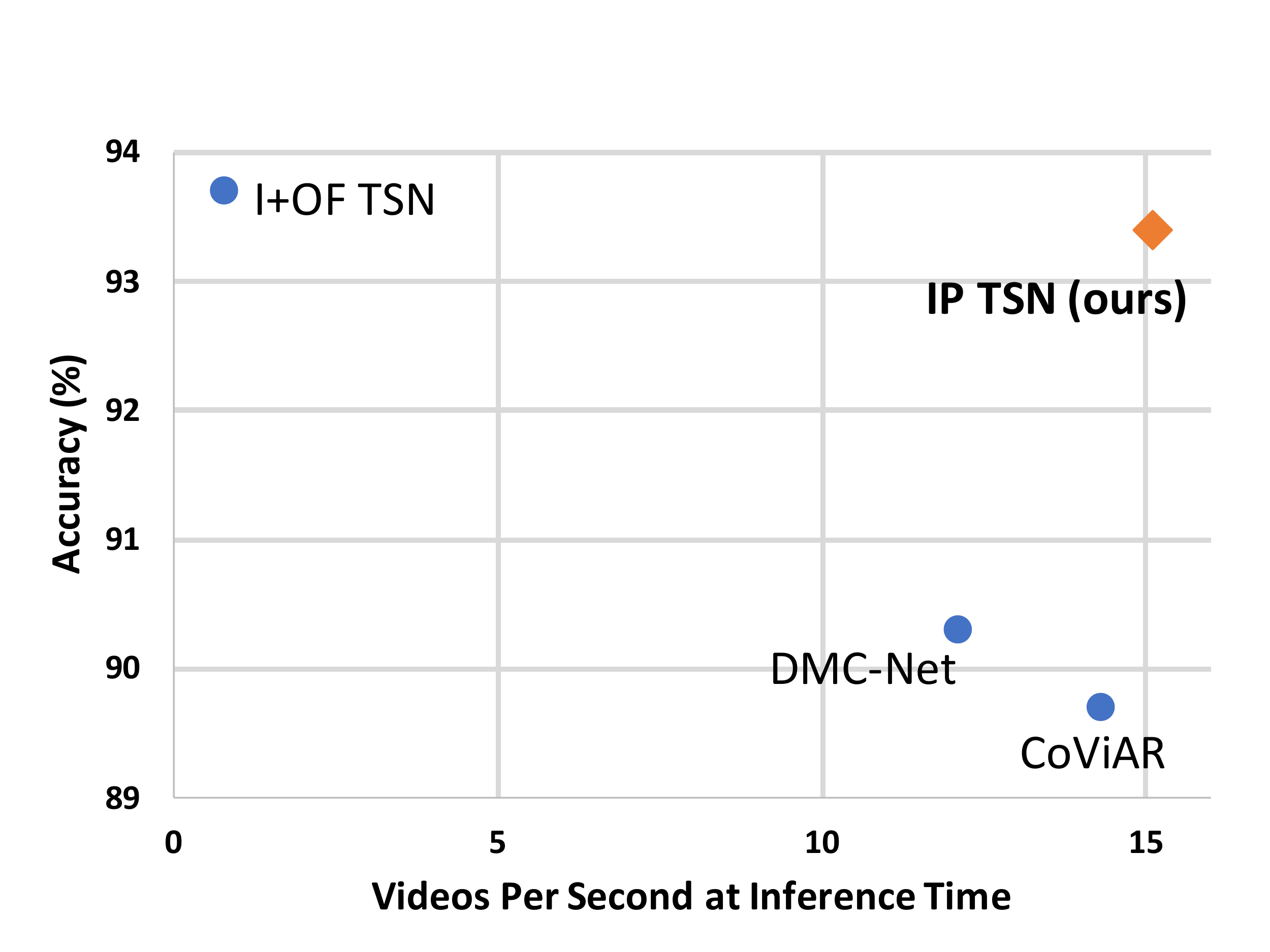}
        \caption{UCF101}
        \label{fig:acc_ucf}
    \end{subfigure}
    \hfill
    \begin{subfigure}[h]{0.3\linewidth}
        \centering
        \includegraphics[width=\linewidth]{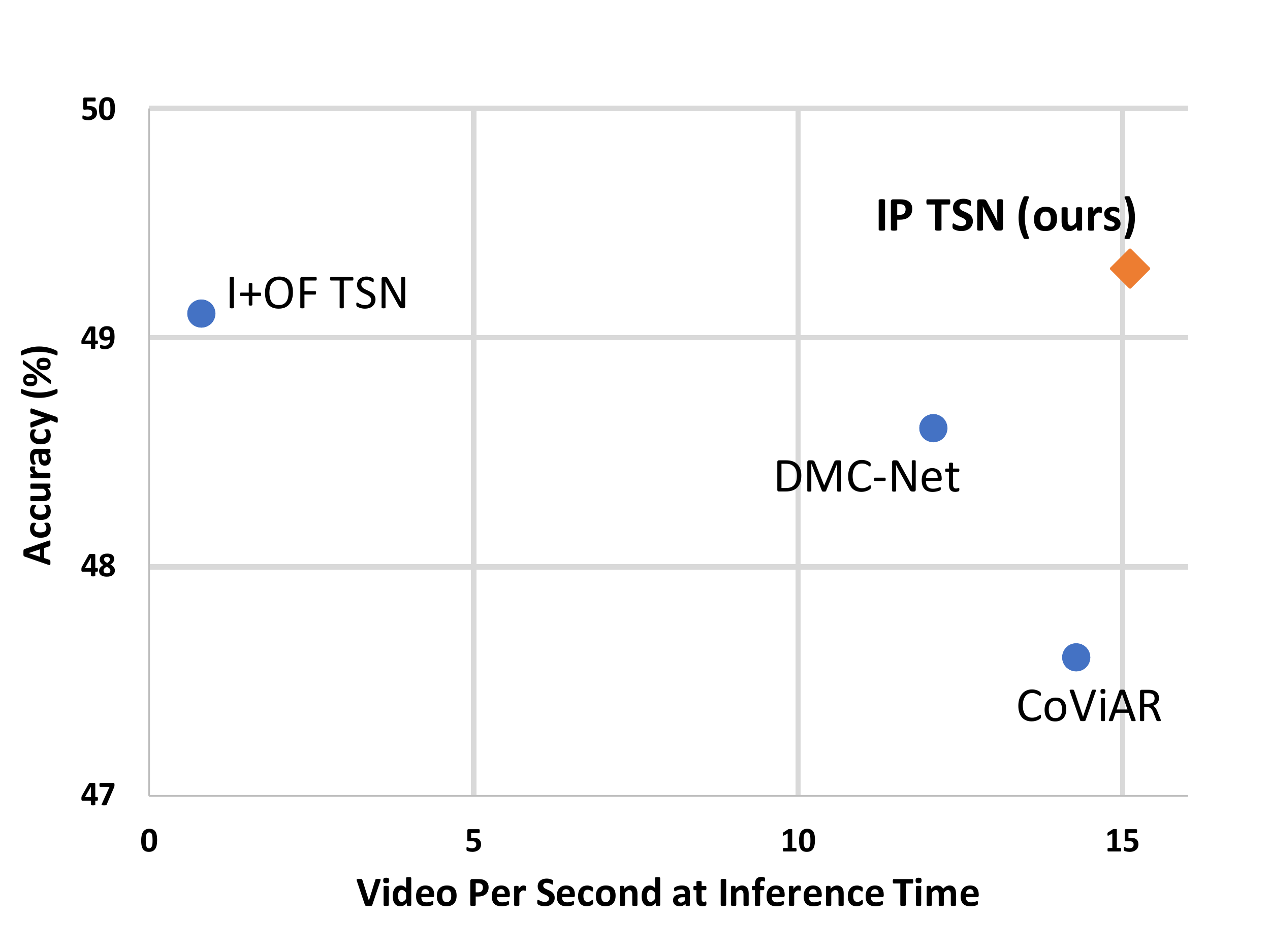}
        \caption{Kinetics-n50}
        \label{fig:acc_kin}
    \end{subfigure}
        \caption{Accuracy vs Videos Per Second at inference time. Input of each network is 16-frames. Results on HMDB51 and UCF101 are averaged over 3 splits. Our IP TSN achieves similar accuracy as I+OF TSN while being about 20$\times$ faster, as and as fast as other compressed domain video recognition methods that perform much worse.}
        \label{fig:acc_datasets}
\end{figure*}

\subsection{P-stream Enhancement with Flow Feature}
\label{sec:p_train}

\textit{OF} is computed from pixel movement between consecutive frames and hence is a full-resolution motion representation of fine details compared to \textit{MV} and \textit{R}.  
It is widely used as the input modality of temporal stream and is proved to be effective \cite{TSN,Kinetics}.
DMC-Net proposes to reconstruct OF-like motion cues at full resolution from \textit{MV} and \textit{R}. 
Recent works~\cite{Zhu_2017_CVPR, repflow2019} demonstrate that the discriminative information embedded in OF is more important than the pixel-level accuracy. 
As a result, we propose to use high-level \textit{OF} feature as supervision to better guide \textit{MV} and \textit{R}. 



As shown in Figure~\ref{fig:pstream}, at training time, the P-stream network is trained with a standard cross-entropy loss using the ground truth class action label as well as the 
the feature output of a pre-trained \textit{OF} network as supervision for knowledge distillation.   \textit{OF} network is pretrained on the same dataset with classification loss. Then the \textit{OF} network weights are frozen when training the P-stream network. 
At test time, the OF network is discarded and only the P-stream network (blue part) remains. 
Note that compared to DMC-Net, we do not need extra layers besides the motion stream feature extractor, so our method is more efficient.

We train our P-stream using the combination of the classification loss and the loss between OF and P-stream features:
\begin{equation}
\label{eq:loss_func}
\mathcal{L}_p = \mathcal{L}_{cls} + \lambda \mathcal{L}_d(f_{p}, f_{OF})
\end{equation}
where $\mathcal{L}_{cls}$ is the cross-entropy loss of P-stream, and $\mathcal{L}_d$ is the Euclidean distance between P-stream feature $f_p$ and OF feature $f_{OF}$. $\lambda$ is the feature loss weight. 
We experiment on different choices of $L_d$ function and its weight $\lambda$,
and finally choose $L1$ loss with $\lambda=50$ which gives the best balance between two losses. 
Note that in general, we could have 
\begin{equation}
\label{eq:loss_func_general}
\mathcal{L}_p = \mathcal{L}_{cls} + \lambda_1 \mathcal{L}_d(f_{p}, f_{OF}) + \lambda_2 \mathcal{L}_s(logit_{p}, logit_{OF})
\end{equation}
where $\mathcal{L}_s$ the soft label cross-entropy loss as is widely used in knowledge distillation. 
But we find in Section~\ref{sec:ablation_study} that it actually downgrades the performance. 
Our insight to this is that \textit{MV} and \textit{R} can capture different properties of the video stream from OF. 
Also, \textit{R} has some object boundary information that OF does not capture but are useful for recognition. 
Hence we do not want to simply reproduce the classification output of OF. 
Instead, we only want OF to help extract more useful motion information from \textit{MV} and \textit{R}.


\section{Experiments}
\label{sec:experiments}
In this section we describe the implementation details and present the performance of our framework for the action recognition task. We will show that our \textbf{IP TSN} achieves both high accuracy and high efficiency.
\subsection{Datasets and Metrics}

We run experiments on 3 action recognition benchmarks:

\noindent \textbf{UCF-101}\cite{soomro2012ucf101}, which contains 13320 videos from 101 action categories. 3 training/testing splits are offered. The performance result is averaged over splits. 

\noindent \textbf{HMDB-51}\cite{kuehne2011hmdb}, which contains 6766 videos from 51 action categories. 3 training/testing splits are offered. The performance result is averaged over splits. 

\noindent \textbf{Kinetics-n50}, following DMC-Net, we use a subset of Kinetics-400 dataset \cite{Kinetics} that contains 30 training videos and 20 testing videos from each of the 400 categories. Training videos are sampled from the original training set, and testing videos are sampled from the original validation set.

\noindent \textbf{Metrics}: As video-level single label is provided by all the above datasets, we report top-1 class prediction accuracy. For efficiency measurement, we report GFLOPs and Videos Per Second \cite{Zolfaghari_2018_ECCV} using 16-frame clip unless specified.

\begin{table*}[tb]
\scriptsize
\center
\begin{tabular}{cc|cc|ccccc|cc}
 &  & \multicolumn{2}{c|}{\begin{tabular}[c]{@{}c@{}}Decoded Video \\ (RGB+Flow) \end{tabular}} & \multicolumn{5}{c|}{\begin{tabular}[c]{@{}c@{}}Compressed Video \\ (I, MV, R) \end{tabular}}  & \multicolumn{2}{c}{OF} \\ \cline{3-11} 
  &   & \multirow{2}{*}{TSN~\cite{TSN}} & \multirow{2}{*}{I3D~\cite{Kinetics}} & \multirow{2}{*}{CoViAR~\cite{wu2018coviar}} & \multirow{2}{*}{DMC-Net~\cite{shou2019dmc}} &
 \multirow{2}{*}{\shortstack{\textbf{I-Stream}\\(ours)}} &
 \multirow{2}{*}{\shortstack{\textbf{P-Stream}\\(ours)}} & \multirow{2}{*}{\shortstack{\textbf{IP TSN}\\(ours)}} & \multirow{2}{*}{OF} &  \multirow{2}{*}{I + OF} \\ 
 & & \multicolumn{2}{c|}{} & \multicolumn{5}{c|}{} \\ \hline
\multicolumn{1}{c|}{\multirow{4}{*}{\begin{tabular}[c]{@{}c@{}}Time \\ (ms) \end{tabular}}} & Preprocessing &  1200 & 1200 & 14.7 & 14.7 & 3.3 & 11.4 & 14.7 & 1200 & 1200 \\ 
 \cline{2-11}
\multicolumn{1}{c|}{} & Spatial Stream & 12.2 &45.5 & 55.3& 55.3 & 43.1 & - & 43.1 & - & 43.1\\
\multicolumn{1}{c|}{} & Temporal Stream & 12.2 & 33.7& - & 12.8 & - & 8.5 & 8.5 & 8.5 & 8.5\\ 
 \cline{2-11}
\multicolumn{1}{c|}{} & Total  &  1224.4 & 1279.2& 70 & 82.8 & 46.4 & 8.5& 66.3 \textbf{}& 1208.5 & 1251.6 \\ \hline \hline
\multicolumn{2}{c|}{\multirow{1}{*}{VPS}}   & 0.8& 0.8 & 14.3 & 12.1 & 21.5 & \textbf{50.2} & \textbf{15.1} & 0.8 & 0.8\\
\hline
\hline
\multicolumn{2}{c|}{\multirow{1}{*}{GFLOPs}} & 33 + 33 & 436 + 401 & 243.4 & 275.6 & 185.6 & \textbf{33.9} & \textbf{219.5} & 185.6 & 33.9 \\
\hline
\hline
\multicolumn{1}{c|}{\multirow{3}{*}{Accuracy (\%)}} & HMDB-51 & 68.5 & 80.7 & 56.5 & 59.0 & 51.5 & 61.8 & \textbf{68.7} & 60.0 & 69.2\\
 \cline{2-11}
\multicolumn{1}{c|}{} & UCF-101 & 94.0 & 98.0 & 89.7 & 90.3 & 86.7 & 87.1 & \textbf{93.4} & 84.9 & 93.7 \\
\cline{2-11}
\multicolumn{1}{c|}{} & Kinetics-n50 & - & - & 48.1 & 48.6 & 45.7 & 23.2 & \textbf{49.6} & 19.1 & 49.3 \\



\end{tabular}
\caption{Comparison of efficiency and accuracy among two stream methods. All compressed domain networks take 16-frame input with one crop. TSN~\cite{TSN} and I3D~\cite{Kinetics} are test as their default settings. 
}
\label{table:speed}
\end{table*}

\subsection{Implementation}

We here detail our training and testing parameters and procedures.

\subsubsection {Training}
We follow the same setting as CoViAR and DMC-Net where videos are resized to $340\times256$ and we use $224\times224$ random cropping and random flipping for data augmentation. 
For I-stream, we employ a ResNet152 classifier and train it with cross-entropy classification loss exactly as CoViAR and DMC-Net. 
The P-stream network is a combination of 2D and 3D CNNs. CoViAR and DMC-Net use ResNet18~\cite{He2015DeepRL} for P-frame modalities. For better comparison and efficiency, we use the same backbone for P-stream. We take the layers of ResNet18 up to conv3x layer for 2D Net, where the output feature has $28\times28$ spatial dimension. For 3D Net, we take 3D-ResNet18~\cite{Zolfaghari_2018_ECCV,tran2017convnet} from conv4x until the end. 
In effect, we replace the later half of the ResNet18 layers with its 3D version. Hence our P-stream is a 2D-3D network. 

In practice, we first train a 2D ResNet18 taking the stacked \textit{MV} and \textit{R} (MR) as input with classification loss only.  We similarly train our \textit{OF} teacher network as a 2D ResNet18 using OF as input with cross-entropy loss. Then, we perform a first round of distillation of the OF features.
Then, we initialize the 2D-3D Res18 with the MR 2D-ResNet18 weights, where the 2D part simply copies the weights and the 3D part inflate the weights from the corresponding layers and we initialize similarly the OF teacher 2D-3D Res18 with copy and inflation from the OF 2D ResNet18 network weights. We then train both 2D-3D networks with cross-entropy loss. 
Finally, the P-stream network is initialized with the MR 2D-3D Res18 and trained with both cross-entropy loss and L1-loss with the average-pooled features of the frozen teacher network as discussed in Section~\ref{sec:p_train}.





\subsubsection{Inference}

For the I-stream, we follow the same setting in CoViAR and DMC-Net except that we uniformly sample 16 I-frames rather than 25. CoViAR and DMC-Net uses 10 $224\times224$ crops for each frame whose scores are averaged to get final I-stream class scores. 
This could introduce a lot of computations. 
We actually find minimal change with only one center-crop as is shown in Table~\ref{table:param_search}. 
For the P-stream, we uniformly sample 16 P-frames using a central $224\times224$ crop from \textit{MV} and \textit{R} as DMC-Net. 
These 16 P-frames form one clip input and is fed into the network to get one P-stream class scores. 
The final prediction is calculated through late fusion of the I-stream and P-stream scores.


\subsection{IP TSN performance}
\label{sec:exp_effective}

Figure~\ref{fig:acc_datasets} shows the accuracy speed trade-off on the three datasets with our method compared to a TSN using I and OF as input and DMC-Net and CoViAR. The numbers are reported on 16-frame input each with one central crop.
Our method clearly exhibits the best trade-off with high speed and high accuracy. 
We also additionally compare with decoded video methods on the left side of Table~\ref{table:speed} and give detailed timings.
Since our P-stream network requires clip input, we measure the inference time per video~\cite{Zolfaghari_2018_ECCV}. 
CoViAR and DMC-Net require 25 frames as input, while ours need 16. 
For fair comparison, we also test CoViAR and DMC-Net with 16 frame-input. 
Speed measurements are performed using one NVIDIA TITAN RTX GPU with CUDA 10 and cuDNN 7.5.0. 
To measure the inference speed for CoViAR and DMC-Net, as well as I-stream, we take a batch of 16 frames and feed into the 2D Nets. 
To measure the inference speed for P-stream, we take one clip of 16 frames as one batch. 
Each frame is centered-cropped without any flipping. The reported results of TSN~\cite{TSN} and I3D~\cite{Kinetics} are based on their default settings, respectively.
As Table~\ref{table:speed} shows, our IP TSN is both faster and more accurate than state-of-the-art compressed domain action recognition methods.  Specifically, our P-stream (50.2VPS) runs more than 60x faster than OF (0.8VPS). The full IP TSN runs about 18x faster than using OF instead. Compared to other compressed domain methods, our IP TSN significantly improves the accuracy while decreasing the GLOPs and increasing VPS both by roughly $20\%$.


\subsection{Ablation study}
\label{sec:ablation_study}
\noindent \textbf{Flow Feature Supervision.} We evaluate our flow feature supervision in Table \ref{tab:distill_effect} on both the original CoViAR based two-stream setting and our IP two-stream setting to show that flow feature supervision enhances the motion information captured from \textit{MV} and \textit{R}.
(1). \textbf{CoViAR TSN} refers to the two-stream methods where separate networks for I, \textit{MV}, \textit{R} are fused with another temporal-stream network for prediction. 
\textbf{CoViAR+MR} trains the \textit{MV}-\textit{R} stacked temporal stream with cross-entropy loss only. \textbf{CoViAR+DMC} is the work proposed in DMC-Net where a generator is used to reconstruct OF-like motion cues from MR and then feed into classifier. \textbf{CoViAR+MR w/ OF-T} uses flow feature as extra supervision in addition to ground truth labels. 
We can see from the table that adding flow supervision does help improve the network ability to extract useful motion information. 
Flow feature supervision performs better than the pixel-level supervision in DMC, while removing the additional generator and reconstruction module. 
(2). \textbf{IP TSN} refers to our IP two-stream network which only consists of a I-stream of \textit{I} and a P-stream network of stacked \textit{MV} and \textit{R} (\textit{MR}). 
We can see that the model trainined with flow feature  supervision (\textbf{I + MR w/ OF-T}) again significantly improves the accuracy over the model trained without (\textbf{I + MR}), which validates the effectiveness of the flow feature distillation.

\noindent \textbf{Flow Distillation Loss.} 
We study commonly used variants in knowledge distillation: L1 loss on last averaged pooled feature (the adopted choice), soft label cross-entropy loss with temperature = 8, and the combination of both which can be exploited jointly in the general loss function introduced in eq.~(\ref{eq:loss_func_general}). 
As shown in Table~\ref{table:feat_variants}, the soft label cross-entropy alone performs worst and actually hurts the performance of the L1 loss when combined with it. 
This may indicate that the L1 loss on features induces a better complementarity with the cross-entropy loss on the ground truth labels than the soft-label cross-entropy loss.
Hence we use only the last pooled feature as supervision  with L1 loss in all other experiments.

\begin{table}[tb]
\footnotesize
\center
\begin{tabular}{cc|c|c}
& & HMDB-51 & UCF-101 \\  \hline
\multicolumn{1}{c|} {\multirow{3}{*}{CoViAR TSN }} & CoViAR+MR &   59.1 &  90.2 \\
\cline{2-4}
\multicolumn{1}{c|}{}  & CoViAR+DMC & 61.8 & 90.9 \\
\cline{2-4}
\multicolumn{1}{c|}{\multirow{1}{*}{ (2D Res18)}}& \textbf{CoViAR+MR w/ OF-T} & \textbf{62.1} & \textbf{91.1} \\
\hline \hline
 \multicolumn{1}{c|}{\multirow{1}{*}{IP TSN }} &  I + MR &  64.9 & 92.5\\
 \cline{2-4}
\multicolumn{1}{c|}{\multirow{1}{*}{(2D-3D Res18) }} &  \textbf{I + MR w/ OF-T} &  \textbf{69.1} &  \textbf{93.4 }\\
\end{tabular}
\caption{Top-1 Accuracy on HMDB-51 and UCF-101 w/ and w/o OF feature supervision (\textit{OF-T}) under two TSN settings. Result is averaged over three splits. 
In both TSN settings, \textit{OF} feature supervision improves the accuracy. It also outperforms DMC which requires extra layers.}
\label{tab:distill_effect}
\end{table}

\begin{table}[tb]
\footnotesize
\center
\begin{tabular}{c|c|c}
 L1 weight $\lambda_1$ & Soft Label Cross-Entropy weight $\lambda_2$ & Top1 Accuracy \\  \hline \hline
0 & 50  & 66.0 \\
 \hline 
25 &  50 & 67.1  \\
 \hline
25 &  25 & 68  \\
 \hline
 50 &  50  & 68  \\
 \hline
\textbf{50} & \textbf{0 }& \textbf{69.1} \\
\end{tabular}
\caption{Top-1 Accuracy on HMDB51 with different training losses. Feature distance gives the best result, as it probably induces a better complementarity with the cross-entropy loss with labels than the soft-label cross-entropy loss. The result is averaged over splits.}
\label{table:feat_variants}
\end{table}

\noindent \textbf{OF vs P-stream} We report on the right side of  Table~\ref{table:speed} the performance of the spatial I-stream, and of the temporal stream either OF or our P-stream or separately. 
We also report the performance of two-streams network with either P-stream or OF as the motion stream. P-stream is roughly \textit{60x} faster than OF while enjoying a higher accuracy.  When combined with I-stream, P-stream is $<0.5\%$  lower than OF-stream.  
We can observe that the OF stream pre-processing time is prohibitive and does not come at any accuracy benefit compared to our P-stream.

\noindent \textbf{Layer inflation.} 
We report in Table~\ref{table:inflate_layer} how the position where we inflate the network affects both the performance and computational cost.  
Note that 'conv1.x' is equivalent to 3D Res18 and  'conv5.x' is equivalent to 2D Res18, i.e., MR 2D-ResNet. We choose to inflate at 'conv3.x' as it gives the  best accuracy-efficiency trade-off.

\begin{table}[tb]
\footnotesize
\center
\begin{tabular}{c|c|c|c|c|c}
  &conv1.x & conv2.x & conv3.x & conv4.x & conv5.x \\  \hline \hline
Top 1 Acc & 61.4 & 70.2 & 69.6 & 66.5 & 62.1\\
\hline
GFLOPs  & 86 & 50 & 34 & 35 & 29
\end{tabular}
\caption{Accuracy on HMDB51 split3 and GLOPs at different inflation layer positions. We choose to inflate at conv3.x as it gives the best accuracy-complexity tradeoff.}
\label{table:inflate_layer}
\end{table}

\begin{figure}[tb]
     \centering
     \begin{subfigure}[h]{0.95\linewidth}
         \centering
         \includegraphics[width=0.95\linewidth]{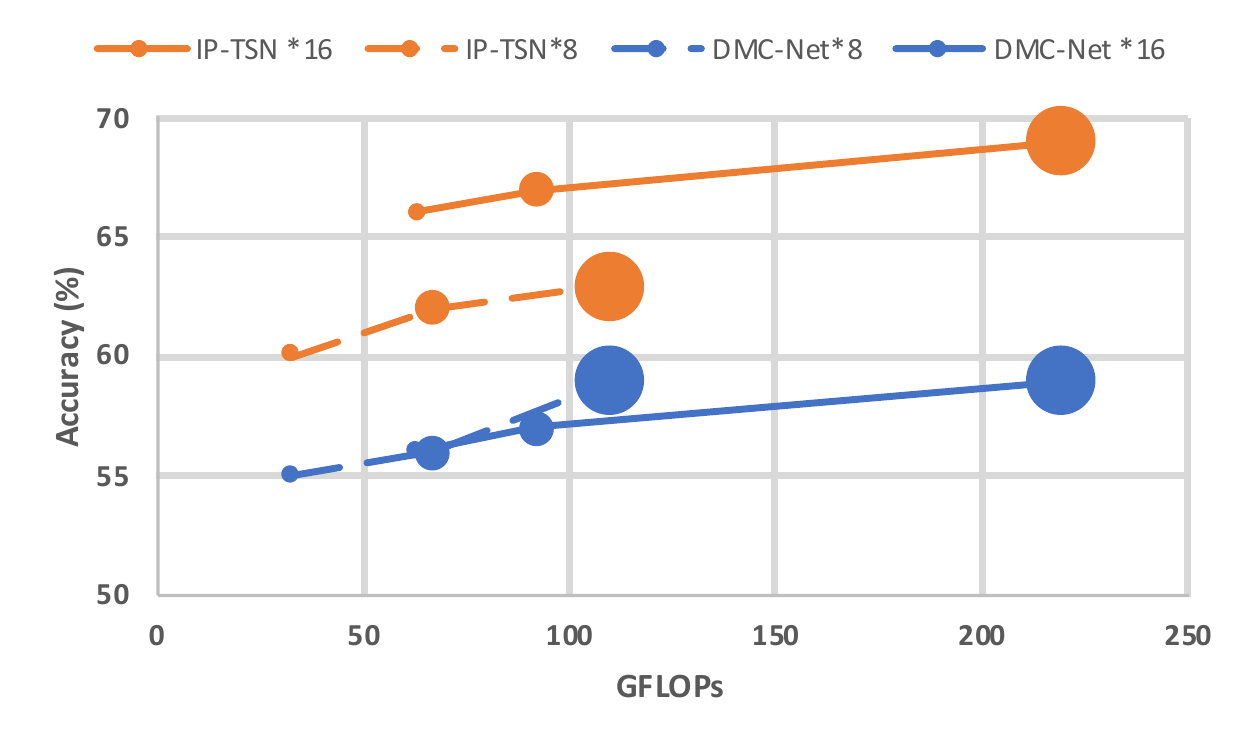}
         \caption{HMDB51}
         \label{fig:stats_hmdb}
     \end{subfigure}
     \hfill
     \begin{subfigure}[h]{0.95\linewidth}
         \centering
         \includegraphics[width=0.95\linewidth]{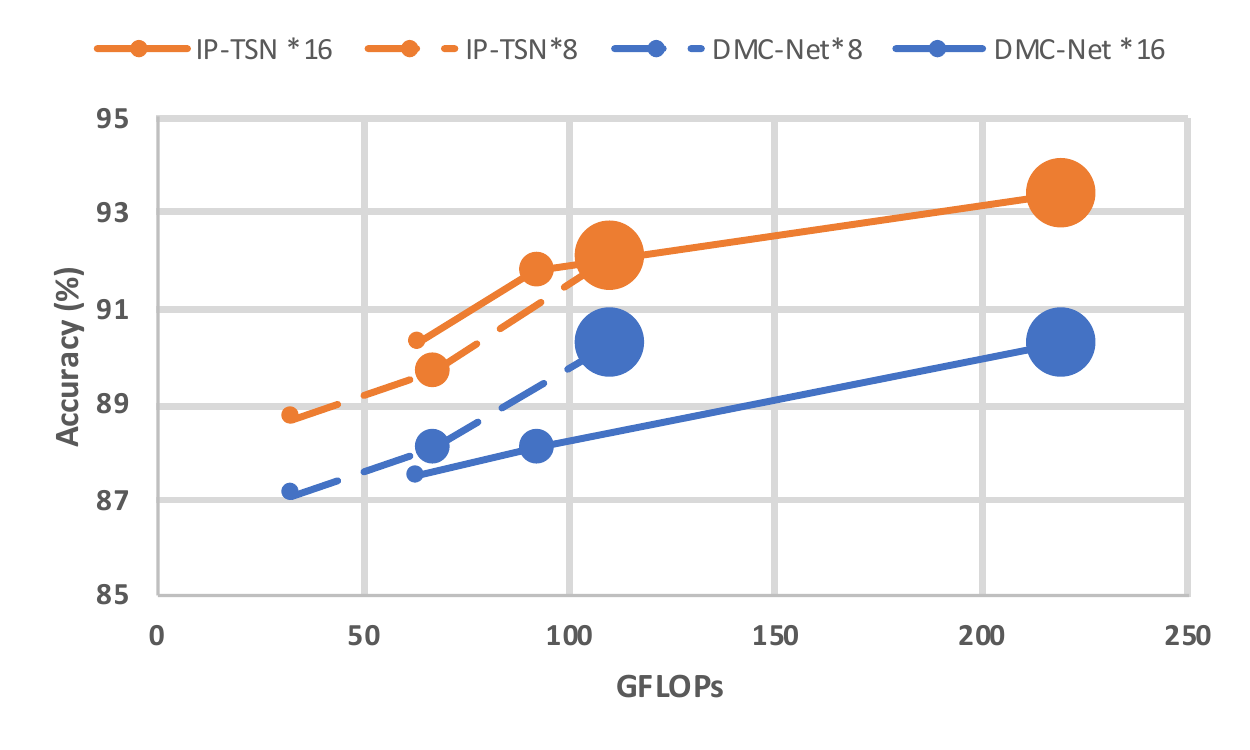}
         \caption{UCF101}
         \label{fig:stats__ucf}
     \end{subfigure}
        \caption{Accuracy vs GFLOPs at inference time. Results on are averaged over 3 splits. This shows how testing segments and I-stream backbone affects the result. Three node sizes corresponds to ResNet18, 34, 152 respectively. Solid and dashed line refer to testing with 16-frames and 8-frames respectively. Orange nodes are result on IP-TSN, and blue nodes are on DMC-Net.}
        \label{fig:stats_net}
\end{figure}

\noindent \textbf{I-Stream Architecture and Input.}
Table 
\ref{table:param_search} gives the Top-1 accuracy on HMDB51 with different I-stream architecture and input sampling strategy. We consider both 8-frame and 16-frame input with one center-crop  or 10 crops. Note that we fix the P-stream architecture as 2D-3D Net we have proposed, and P-stream does not take multi-crop inputs. We can see from the table that 2D-3D Res18 does not help with I-stream. One reason for this is that I-frame are rare in short videos hence hardly provide any useful spatiotemporal information. 
Increasing crops significantly adds to the complexity without significant improvement, while clip length affect substantially. 
However, increasing the clip length improves the performance substantially. We get our best results with ResNet152 as is also used in DMC-Net and CoViAR. This validates our design concept that IP streams should use non-symmetric networks. We also report in Figure~\ref{fig:stats_net} the accuracy as a function of GFLOPs obtained by our method and the DMC-Net approach using different I-stream network architectures: ResNet18, ResNet34, and ResNet152, and how performance changes with either 8 or 16 frames as input for each video.
We see that at a same number of GFLOPs our method always outperforms DMC-Net.

\begin{table}[tb]
\scriptsize
\center
\begin{tabular}{c|cc|cc|c|cc}
 \multicolumn{1}{c|}{\multirow{2}{*} {\begin{tabular}[c]{@{}c@{}} I frame \\ backbone \end{tabular}}}  & \multicolumn{2}{c|}{Istream}&\multicolumn{2}{c|}{Pstream}&\multicolumn{1}{c|}{\multirow{2}{*} {\begin{tabular}[c]{@{}c@{}} VPS  \\ {} \end{tabular}}}  & \multicolumn{1}{c}{Top-1 Acc.}   \\ 
 \cline{2-5}
 & seg*crop& Acc. & seg*crop & Acc. & & HMDB51  \\
 \hline
  \multicolumn{1}{c|}{2D-3D Res18} 
   & 16*1 & - & 16*1 & 61.9 & - & 66.8\\ 
 \hline
  \multicolumn{1}{c|}{\multirow{4}{*} {ResNet 18}} 
   & 8*1 &44.1 & \multicolumn{1}{c}{\multirow{2}{*} {8*1}} & \multicolumn{1}{c|}{\multirow{2}{*}{53.1}} & 57.7 & 60.1\\ 
 & 8*10 & 44.6 & & &26.5 & 60.2\\
 \cline{2-7}
 & 16*1 & 44.3 & \multicolumn{1}{c}{\multirow{2}{*} {16*1}} & \multicolumn{1}{c|}{\multirow{2}{*}{61.9}} & 52.1  &66.3 \\ 
 & 16*10 & 45 & & &13.7 & 66.2\\
 \hline 
  \multicolumn{1}{c|}{\multirow{4}{*} {ResNet 152}}   & 8*1 & 51.1 & \multicolumn{1}{c}{\multirow{2}{*} {8*1}} & \multicolumn{1}{c|}{\multirow{2}{*}{53.1}} & 24  &63.4\\ 
 & 8*10 & 51.4 & & & 4.7 & 63.6\\
 \cline{2-7}
 & 16*1 &51.5  & \multicolumn{1}{c}{\multirow{2}{*} {16*1}} & \multicolumn{1}{c|}{\multirow{2}{*}{61.9}} & 15.1 & \textbf{68.7}\\ 
 & 16*10 & 51.8 & & & 2.4 & \textbf{69.1}\\

\end{tabular}
\caption{Accuracy on HMDB51 using different I-frame backbones and number of segments and crops. The result is averaged across 3 splits. 
}
\label{table:param_search}
\end{table}


\subsection{Comparison with State-of-the-Art}

We finally compare our IP two-stream network with current state-of-the-art methods in Table~\ref{table:res_decoded}. Our method outperforms all other compressed-domain methods significantly by $7\%$ and $2.5\%$ on HMDB51 and UCF101. We have brought the compressed-domain methods performance significantly closer to the decoded video methods. Without pre-training on large-scalre video datasets, we are competitive with many decoded methods, apart from the most expensive two-stream methods using heavy architectures such as I3D. For example, our proposed IP TSN even achieves better performance on both HMDB-51 and UCF-101 than ECO~\cite{zolfaghari2018eco}, a recently proposed decoded-video based efficient video understanding model, under the same amount of frames (16) to process. Note that as is shown in Table \ref{table:speed}, all the decoded two stream based networks need to compute optical flow, which is often the main bottleneck of real time processing. However, we are able to greatly speed up the process while maintaining a comparable performance.

\begin{table}[tb]
\footnotesize

\center
\begin{tabular}{lcc}
\multicolumn{1}{c}{} & \multicolumn{1}{c}{HMDB-51} & \multicolumn{1}{c}{UCF-101} \\ \hline
\multicolumn{3}{l}{\textbf{Compressed video based methods}} \\
 &  &  \\
EMV-CNN \cite{ZhangWWQW16_MVCNN} & 51.2 (split1) & 86.4 \\
DTMV-CNN \cite{Zhang2018RealTimeAR} & 55.3 & 87.5 \\
CoViAR \cite{wu2018coviar} & 59.1 & 90.4 \\
DMC-Net \cite{shou2019dmc} & 61.8 & 90.9 \\ 
\textbf{CoViAR+MR w/ OF-T [ours]} & \textbf{62.1} & \textbf{91.1} \\
\textbf{IP TSN [ours]} & \textbf{69.1} & \textbf{93.4}\\ \hline
\multicolumn{3}{l}{\textbf{Decoded video based methods} \textbf{\textit{(RGB only)}}} \\
\multicolumn{2}{c}{\textit{\textbf{Frame-level classification}}} & \\
ResNet-50 \cite{He_2016_CVPR} & 48.9 & 82.3 \\
ResNet-152 \cite{He_2016_CVPR} & 46.7 & 83.4 \\
\multicolumn{2}{c}{\textit{\textbf{Motion representation learning}}} & \\
ActionFlowNet \cite{ng2018actionflownet} & 56.4 & 83.9 \\ 
PWC-Net (ResNet-18) + CoViAR \cite{Sun2018PWC-Net} & 62.2 & 90.6 \\
TVNet \cite{fan2018end} & 71.0 & 94.5 \\
\multicolumn{2}{c}{\textit{\textbf{Efficient spatio-temporal modeling}}} & \\
 ECO$_{16f}$~\cite{zolfaghari2018eco} & 68.5 & 92.8\\
 ECO$_{En}$~\cite{zolfaghari2018eco} & 72.4 & 94.8\\
 \hline
\multicolumn{3}{l}{\textbf{Decoded video based Two Stream methods} \textbf{\textit{(RGB + Flow)}}} \\
Two-stream \cite{Simonyan14b} & 59.4 & 88.0 \\
Two-Stream fusion \cite{feichtenhofer2016convolutional} & 65.4 & 92.5 \\
I3D \cite{Kinetics} & 80.7 & 98.0 \\
R(2+1)D \cite{tran2018closer} & 78.7 & 97.3
\end{tabular}
\caption{Accuracy averaged over all three splits on HMDB-51 and UCF-101 for both state-of-the-art compressed video based methods and decoded video based methods.}
\label{table:res_decoded}
\end{table}


\section{Conclusion}
In this work, we propose a new IP two-stream framework for compressed video action recognition, where the different frames (I and P) of the compressed representation are exploited for different modeling purposes (spatial and motion, respectively). 
We also propose an efficient P-stream training strategy using an optical flow feature supervision to train a 2D-3D architecture for the motion stream. Our P-stream is 60$\times$ faster than an \textit{OF} stream with similar accuracy.
Evaluation on public benchmarks validate our method and shows significant improvement over prior compressed domain action recognition method and approach decoded video methods performance. 
Overall our method is highly efficient as it can process 15 videos per second at a high level of accuracy. 

\newpage
{\small
\bibliographystyle{ieee_fullname}
\bibliography{egbib}
}

\end{document}